\documentclass[11pt]{article}
\usepackage[margin=1in]{geometry}
\usepackage{amsmath}
\usepackage{amsthm}
\usepackage{graphicx}
\usepackage[breaklinks,hidelinks]{hyperref}
\usepackage{accents}
\usepackage[authoryear]{natbib}
\newcommand{\E}{{\bf E}}

\newlength{\dhatheight}

\newlength{\dtildeheight}
\newcommand{\doubletilde}[1]{{%
 \settoheight{\dtildeheight}{\ensuremath{\tilde{#1}}}%
 \addtolength{\dtildeheight}{-0.16ex}%
 \tilde{\vphantom{\rule{1pt}{\dtildeheight}}%
 \smash{\tilde{#1}}}%
}}

\newtheorem{theorem}{Theorem}[]

\newtheorem{remark1}[theorem]{Remark}
\newenvironment{remark}{\begin{remark1} \rm}{\end{remark1}}

\newcommand{\captionfonts}{\normalsize}

\makeatletter  
\long\def\@makecaption#1#2{%
  \vskip\abovecaptionskip
  \sbox\@tempboxa{{\captionfonts #1: #2}}%
  \ifdim \wd\@tempboxa >\hsize
    {\captionfonts #1: #2\par}
  \else
    \hbox to\hsize{\hfil\box\@tempboxa\hfil}%
  \fi
  \vskip\belowcaptionskip}
\makeatother   

\begin{document}


\centerline{\LARGE A Mathematical Motivation for}
\vspace{.075in}
\centerline{\LARGE Complex-valued Convolutional Networks}
\vspace{.05in}

\ \\
{\large Joan Bruna,}
{\large Soumith Chintala,}
{\large Yann LeCun,}
{\large Serkan Piantino,}
{\large Arthur Szlam,}
{\large Mark Tygert}\\\ \\
{Facebook Artificial Intelligence Research,
1 Facebook Way, Menlo Park, California 94025}\\
%


\markboth{}{NC instructions}
\ \\
%
Abstract:\\ A complex-valued convolutional network (convnet)
implements the repeated application of the following composition
of three operations, recursively applying the composition
to an input vector of nonnegative real numbers:
(1) convolution with complex-valued vectors
followed by (2) taking the absolute value of every entry
of the resulting vectors followed by (3) local averaging.
For processing real-valued random vectors, complex-valued convnets
can be viewed as ``data-driven multiscale windowed power spectra,''
``data-driven multiscale windowed absolute spectra,''
``data-driven multiwavelet absolute values,'' or
(in their most general configuration)
``data-driven nonlinear multiwavelet packets.''
Indeed, complex-valued convnets can calculate multiscale windowed spectra
when the convnet filters are windowed complex-valued exponentials.
Standard real-valued convnets, using rectified linear units (ReLUs),
sigmoidal (for example, logistic or tanh) nonlinearities,
max.\ pooling, etc., do not obviously exhibit the same exact correspondence
with data-driven wavelets (whereas for complex-valued convnets,
the correspondence is much more than just a vague analogy).
Courtesy of the exact correspondence, the remarkably rich and rigorous body
of mathematical analysis for wavelets applies directly
to (complex-valued) convnets.

\section{Introduction}

Convolutional networks (convnets) have become increasingly important
to artificial intelligence in recent years,
as reviewed by~\citet{lecun-bengio-hinton}.
The present paper presents a theoretical argument for complex-valued convnets
and their remarkable performance; complex-valued convnets turn out to calculate
``data-driven multiscale windowed spectra'' characterizing
certain stochastic processes common in the modeling
of time series (such as audio) and natural images
(including patterns and textures).
We motivate the construction of such multiscale spectra
via ``local averages of multiwavelet absolute values''
or, more generally, ``nonlinear multiwavelet packets.''

A textbook treatment of all concepts and terms used above and below
is given by~\citet{mallat2008}.
Further information is available in the original work of~\citet{daubechies},
\citet{meyer}, \citet{coifman-meyer-quake-wickerhauser},
\citet{coifman-donoho}, \citet{simoncelli-freeman}, \citet{meyer-coifman},
\citet{lecun-bottou-bengio-haffner},
\citet{donoho-mallat-von_sachs-samuelides},
\citet{srivastava-lee-simoncelli-zhu}, \citet{rabiner-schafer},
and~\citet{mallat2008}, for example.
The work of~\citet{haensch-hellwich}, \citet{mallat2010},
\citet{poggio-mutch-leibo-rosasco-tacchetti},
\citet{bruna-mallat}, \citet{bruna-mallat-bacry-muzy},
and~\citet{chintala-ranzato-szlam-tian-tygert-zaremba}
also develops complex-valued convnets, providing copious applications
and numerical experiments.
A related, more sophisticated connection (to renormalization group theory)
is given by~\citet{mehta-schwab}.
Our exposition relies on nothing but the basic signal processing
treated by~\citet{mallat2008}. Via the connections discussed below,
the rich, rigorous mathematical analysis surveyed by~\citet{daubechies},
\citet{meyer}, \citet{mallat2008}, and others applies directly
to complex-valued convnets.

Citing such connections, the present paper's anonymous reviews suggested
viewing complex-valued convnets as a kind of baseline architecture for much
of the deep learning reviewed by~\citet{lecun-bengio-hinton}.
Section~\ref{numerical} presents numerical analyses corroborating
this viewpoint.
Having such a theoretical basis for deep learning could help in paring down the
combinatorial explosion of possibilities for future developments, while
probably also illuminating further possibilities.

The present paper proceeds as follows:
Section~\ref{stationary} reviews stationary stochastic processes
and their spectra.
Section~\ref{localstat} reviews locally stationary stochastic processes
and the connection of their spectra to stages in a complex-valued convnet.
Section~\ref{multiscale} introduces multiscale (multiple stages in a convnet).
Section~\ref{fitting} describes the fitting/learning/training that
the connection to convnets facilitates.
Section~\ref{numerical} briefly compares on a common benchmark the accuracies
for the complex-valued convnets
of~\citet{chintala-ranzato-szlam-tian-tygert-zaremba} to those
for the scattering transforms of~\citet{mallat2010} and
for the standard real-valued convnets of~\citet{krizhevsky-sutskever-hinton}.
Section~\ref{conclusion} generalizes and summarizes
the aforementioned sections.

\section{Stationary stochastic processes}
\label{stationary}

For simplicity, we first limit consideration to the special case
of a doubly infinite sequence of nonnegative random variables $X_k$,
where $k$ ranges over the integers.
This input data will be the result of convolving an unmeasured
independent and identically distributed (i.i.d.)\ sequence
$Z_k$, where $k$ ranges over the integers, with an unknown sequence
of real numbers $f_k$, where $k$ ranges over the integers
(this latter sequence is known as a ``filter,''
whereas the i.i.d.\ sequence is known as ``white noise''):
\begin{equation}
\label{colored}
X_j = \sum_{k=-\infty}^{\infty} f_{j-k} \, Z_k
\end{equation}
for any integer $j$.
Such a sequence $X_k$, with $k$ ranging over the integers,
is a (strictly) ``stationary stochastic process.''
The terminology ``strictly stationary'' refers to the fact that lagging
or shifting the process preserves the probability distribution of the process:
indeed, for any integer $l$, the shift $Y_k = X_{k-l}$, where $k$ ranges
over the integers, satisfies
\begin{equation}
Y_j = \sum_{k=-\infty}^{\infty} f_{j-k} \, Z'_k
\end{equation}
for any integer $j$, where $Z'_k = Z_{k-l}$;
the sequence $Z'_k$, with $k$ ranging over the integers, is i.i.d.\ 
with the same distribution as $Z_k$, where $k$ ranges over the integers.

The associated ``absolute spectrum'' is
\begin{equation}
\label{absspec}
\tilde{X}(\omega) = \lim_{n \to \infty} \E\left|
\frac{1}{\sqrt{2n+1}} \sum_{k=-n}^n e^{-i k \omega } X_k \right|
\end{equation}
for any real number $\omega$ (usually we consider not just any, but instead
restrict consideration to a sequence running from $0$ to about $2\pi$).
Please note that lagging or shifting the process changes neither
the probability distribution of the process (since the process is stationary)
nor the absolute spectrum:
for any integer $l$, the shift $Y_k = X_{k-l}$
yields $\tilde{Y}(\omega) = \tilde{X}(\omega)$ for any real number $\omega$,
due to the absolute value in equation~\ref{absspec}.

Similarly, the associated ``power spectrum'' is
\begin{equation}
\label{powerspec}
\doubletilde{X}(\omega) = \lim_{n \to \infty} \E\left(\left|
\frac{1}{\sqrt{2n+1}} \sum_{k=-n}^n e^{-i k \omega } X_k \right|^2\right)
\end{equation}
for any real number $\omega$; there is an extra squaring under the expectation
in equation~\ref{powerspec} compared to equation~\ref{absspec}.
Again, lagging or shifting the process changes neither
the probability distribution of the process nor the power spectrum:
for any integer $l$, the shift $Y_k = X_{k-l}$
yields $\doubletilde{Y}(\omega) = \doubletilde{X}(\omega)$
for any real number $\omega$, due to the absolute value
in equation~\ref{powerspec}.
The remainder of the present paper focuses on the absolute spectrum;
most of the discussion applies to the power spectrum, too.

\begin{remark}
The absolute spectrum can be more robust than the power spectrum,
in the same sense that the mean absolute deviation can be more robust
than the variance or standard deviation. The power spectrum is more fundamental
in a certain sense, yet the absolute spectrum may be preferable
for applications to machine learning. We conjecture that both can work
about the same. We focus on the absolute spectrum to simplify the exposition.
\end{remark}

\section{Locally stationary stochastic processes}
\label{localstat}

In practice, the input data is seldom strictly stationary,
but usually only locally stationary, that is, equation~\ref{colored} becomes
\begin{equation}
\label{slow}
X_j = \sum_{k=-\infty}^{\infty} f^{(j)}_{j-k} \, Z_k
\end{equation}
for any integer $j$, where $f^{(j)}_k$ changes much more slowly
when changing $j$ than when changing $k$.
To accommodate such data, we introduce windowed spectra;
for any even nonnegative-valued sequence $g_k$,
with $k$ ranging through the integers
--- this sequence could be samples of a Gaussian or any other window suitable
for Gabor analysis (the data itself will determine $g$ during training) ---
we consider
\begin{equation}
\label{localized}
\tilde{X}_l(\omega) = \frac{1}{2n+1} \sum_{j=-n+l}^{n+l}
\left| \frac{1}{\sqrt{2n+1}}
\sum_{k=-\infty}^{\infty} e^{-i k \omega } g_{k-j} X_k \right|
\end{equation}
for any integer $l$, with some positive integer $n$.
The extra summation in equation~\ref{localized} averages away noise
and is a kind of approximation to the expected value in equation~\ref{absspec}.
Usually $g_k$ is fairly close to $1$ for $k = -n$, $-n+1$, \dots, $n-1$, $n$,
and $g_k$ is fairly close to $0$ for $|k| > n$, making a reasonably smooth
transition between 0 and 1.
The most important difference between equation~\ref{absspec}
and equation~\ref{localized} is the absence of a limit in the latter
(hence the terminology, ``local'' spectrum).

Due to the absolute value, equation~\ref{localized} is equivalent to
\begin{equation}
\label{conv}
\tilde{X}_l(\omega) = \frac{1}{2n+1} \sum_{j=-n+l}^{n+l}
\left| \frac{1}{\sqrt{2n+1}}
\sum_{k=-\infty}^{\infty} g_{j-k}(\omega) \, X_k \right|
\end{equation}
for any {\it even} nonnegative-valued sequence $g_k$,
with $k$ ranging through the integers, where
\begin{equation}
g_k(\omega) = e^{i k \omega} g_k
\end{equation}
for any integer $k$ (``even'' means that $g_{-k} = g_k$ for every integer $k$).
Please note that the right-hand side of equation~\ref{conv} is just
a convolution followed by the absolute value followed by local averaging;
this will facilitate fitting/learning/training using data ---
enabling a ``data-driven'' approach --- in Section~\ref{fitting}.

\section{Multiscale}
\label{multiscale}

\begin{figure}
\vspace{1in}
\centering
\parbox{\textwidth}{\includegraphics[width=\textwidth]{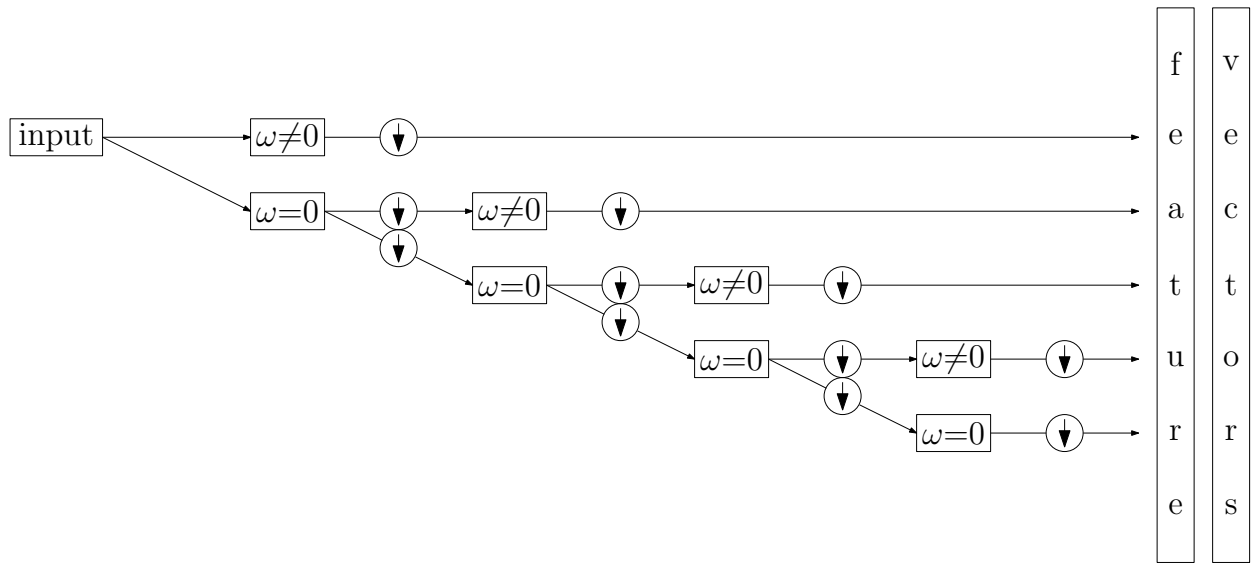}}
\vspace{.25in}
\caption{A flow chart for the ``wavelet transform'' of an input vector:
each box ``$\omega$$=$$0$'' corresponds to equation~\ref{conv}
with $\omega$$=$$0$ or (equivalently) to equation~\ref{convolution};
each box ``$\omega$$\ne$$0$'' corresponds to equation~\ref{conv}
--- convolution followed by taking the absolute value
of every entry followed by local averaging; each circle ``$\downarrow$''
corresponds to subsampling (say, retaining only every other entry)
}
\label{wavefig}
\vspace{1in}
\end{figure}

\begin{figure}
\centering
\parbox{\textwidth}{\includegraphics[width=\textwidth]{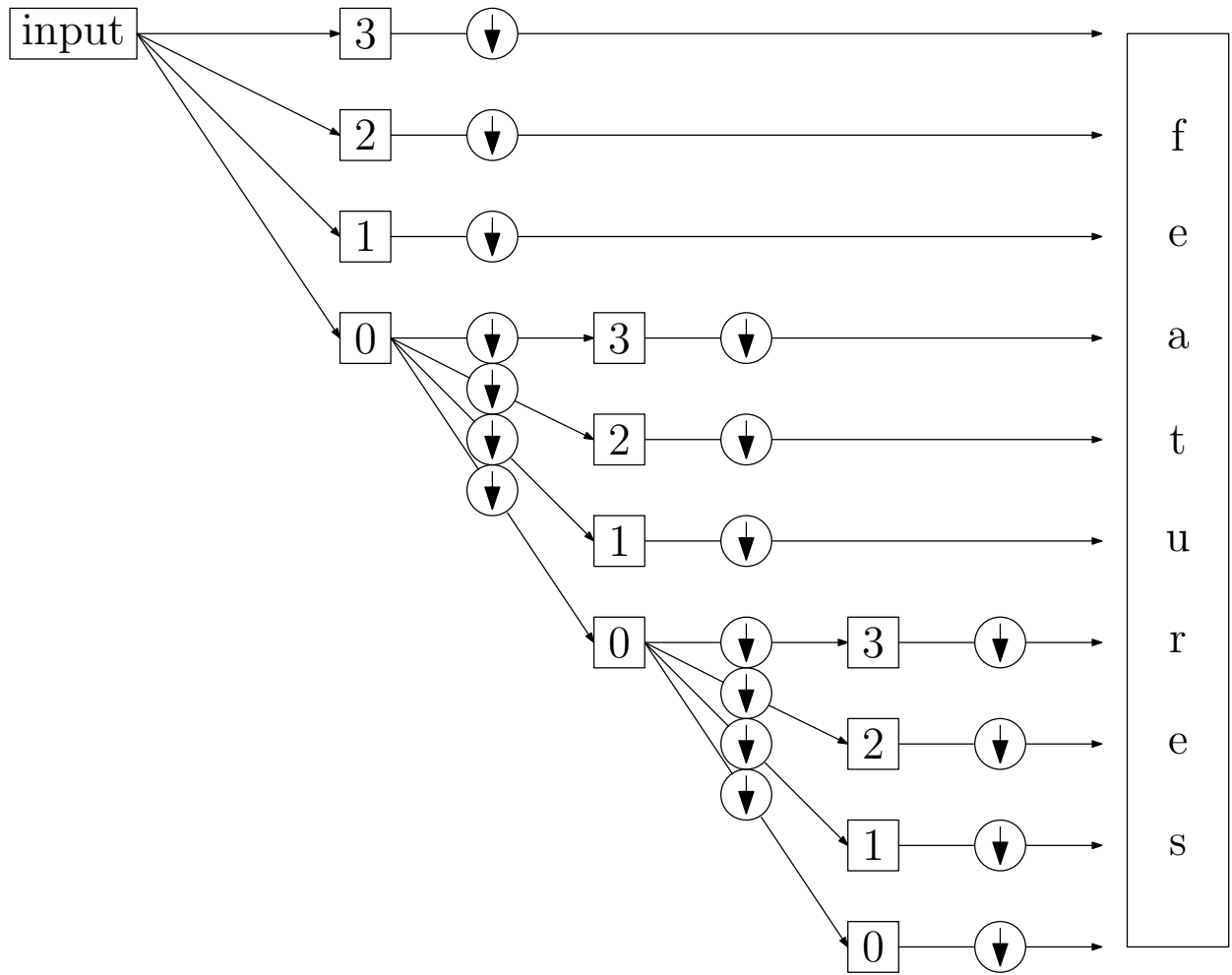}}
\vspace{.25in}
\caption{A flow chart for the ``multiwavelet transform'' of an input vector:
each box ``0'' corresponds to equation~\ref{conv} with $\omega$$=$$0$
or (equivalently) to equation~\ref{convolution};
each box ``1,'' ``2,'' or ``3'' corresponds
to equation~\ref{conv} for different convolutional filters,
but always with convolution followed by taking the absolute value
of every entry followed by local averaging; each circle ``$\downarrow$''
corresponds to subsampling (say, retaining only every fourth entry)
}
\label{multifig}
\end{figure}

\begin{figure}
\centering
\parbox{\textwidth}{\includegraphics[width=\textwidth]{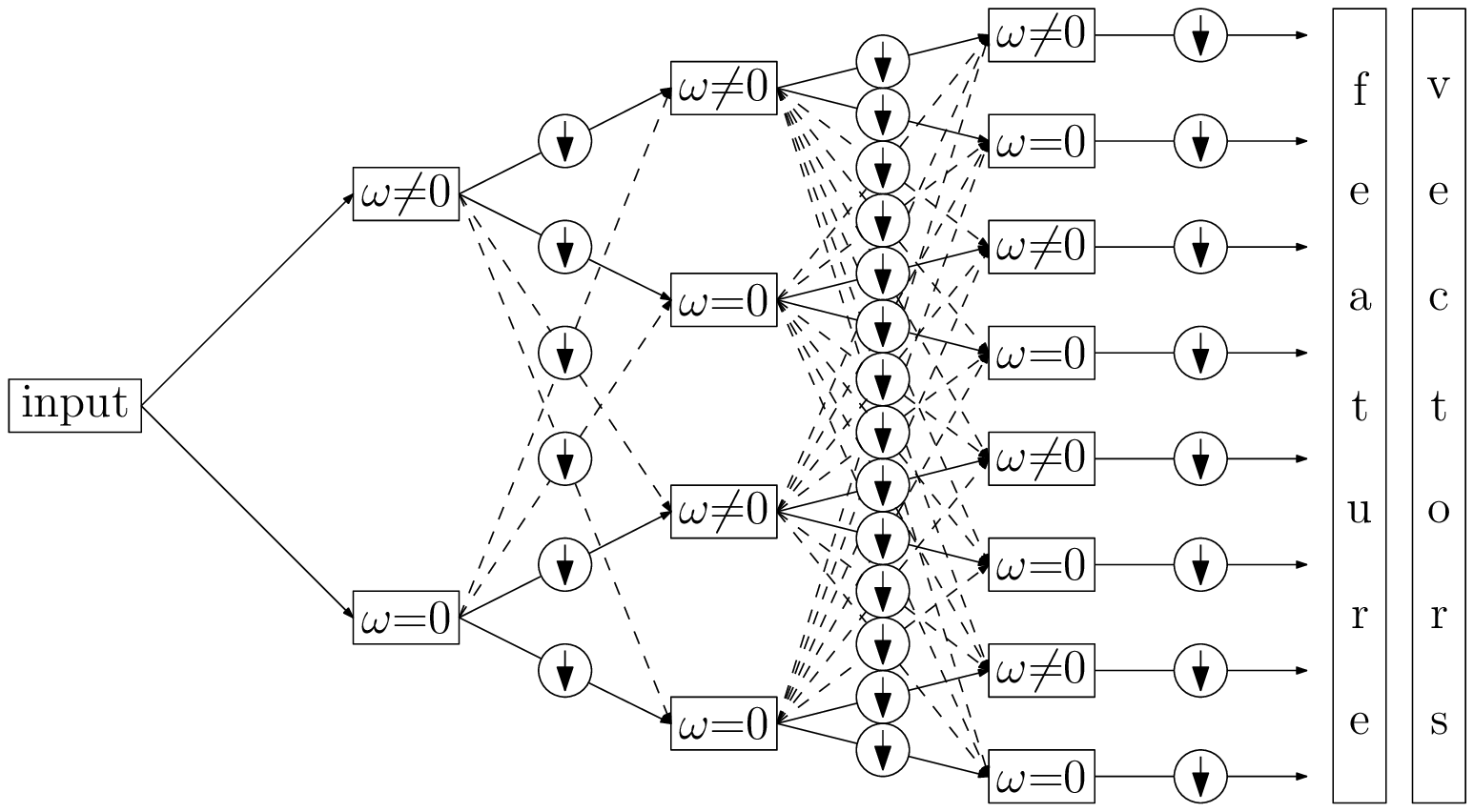}}
\vspace{.25in}
\caption{A flow chart for the ``nonlinear wavelet packet transform''
of an input vector:
each box ``$\omega$$=$$0$'' corresponds to equation~\ref{conv}
with $\omega$$=$$0$ or (equivalently) to equation~\ref{convolution};
each box ``$\omega$$\ne$$0$'' corresponds to equation~\ref{conv}
--- convolution followed by taking the absolute value
of every entry followed by local averaging; each circle ``$\downarrow$''
corresponds to subsampling (say, retaining only every other entry);
the dashed arrows can involve downweighting the associated summands
(and the convolutional filter can be different for every arrow);
Figure~\ref{wavefig} is essentially a special case of the present figure
for which some of the convolutional filters simply deconvolve
the preceding local averaging (omitting some of the subsampling)
}
\label{packetfig}
\end{figure}

In most cases, the ideal choices of $n$ and width of the window
in equation~\ref{conv}, that is, the ideal number of indices for which $g_k$ is
substantially nonzero, are far from obvious. Often, in fact, multiple widths
are relevant (say, wider for lower-frequency variations than
for higher frequency).
Not knowing the ideal a priori, we use multiple windows on multiple scales.
An especially efficient multiscale implementation processes the results
of the lowest-frequency channels recursively.
For the lowest frequency, $\omega = 0$,
and when $X_k$ is nonnegative for every integer $k$
(for example, the input $X_k$ could be the $\tilde{X}_k$
arising from previous processing), equation~\ref{conv} simplifies to
\begin{equation}
\label{convolution}
\tilde{X}_l(0) = \frac{1}{\sqrt{2n+1}} \sum_{k=-\infty}^{\infty} h_{l-k} X_k
\end{equation}
for any integer $l$, where
\begin{equation}
h_l = \frac{1}{2n+1} \sum_{j=-n+l}^{n+l} g_j
\end{equation}
for any integer $l$,
and again $g_j$, with $j$ ranging through the integers,
is an even sequence of nonnegative real numbers
(``even'' means that $g_{-j} = g_j$ for every integer $j$).
The result of equation~\ref{convolution} is simply a convolution
with the input sequence, and further convolutions --- say via recursive
processing of the form in equation~\ref{conv} --- can undo this convolution
and set the effective window however desired in later stages.
The deconvolution and subsequent convolution with the windowed exponential
of a later stage is numerically stable if the later window is wider
than the preceding.
In particular, recursively processing the zero-frequency channels
in this way can implement a ``wavelet transform'' (if each recursive stage
considers only two values for $\omega$, one zero and one nonzero
--- see Figure~\ref{wavefig})
or a ``multiwavelet transform'' (if each recursive stage considers
multiple values for $\omega$, with one of the values being zero
--- see Figure~\ref{multifig}).
For multidimensional signals, multiwavelets detect local directionality
beyond what wavelets provide.
If we recursively process the higher-frequency channels, too,
then we obtain a ``nonlinear wavelet packet transform'' or
a ``nonlinear multiwavelet packet transform'' --- a kind of nonlinear
iterated filter bank --- see Figure~\ref{packetfig}.
Linearly recombining the different frequency channels may help realize
local rotation-invariance and other potentially desirable properties
(indeed, \citet{mallat2010} did this for rotations and other transformations)
--- including generating harmonics when processing audio signals.
The transforms just discussed are undecimated,
but interleaving appropriate decimation or subsampling applied to the sequences
yields the usual decimated transforms.

\begin{remark}
In practice, decimation or subsampling is important
to avoid overfitting in the data-driven approach discussed below,
by limiting the number of degrees of freedom appropriately.
Even when the signal is not a strictly stationary stochastic process,
the averaging in equation~\ref{conv} --- the leftmost summation ---
performs the ``cycle spinning'' of~\citet{coifman-donoho} to avoid artifacts
that would otherwise arise due to windows' partitioning after subsampling.
The averaging reduces the variance; wider averaging would further reduce
the variance.
\end{remark}

\begin{remark}
Sequences that are finite rather than doubly infinite provide only enough
information for estimating a smoothed version of the spectrum. Alternatively,
a finite amount of data provides information for estimating multiscale
windowed spectra yielding time-frequency (or space-Fourier) resolution similar
to the multiresolution analysis of wavelets.
\end{remark}

\begin{remark}
SIFT, HOG, SURF, etc.\
of~\citet{lowe1999}, \citet{lowe2004}, \citet{dalal-triggs},
\citet{bay-ess-tuytelaars-van_gool}, and others are more analogous
to the multiwavelet architecture of Figure~\ref{multifig}
than to the more general wavelet-packet architecture of Figure~\ref{packetfig}.
\end{remark}

\section{Fitting/learning/training}
\label{fitting}

The ``multiwavelet transform'' constitutes a desirable baseline model.
We can easily adapt to the data the choices of windows and indeed the whole
recursive structure of the processing (whether restricting the recursion
to the zero-frequency channels, or also allowing the recursive processing
of higher-frequency channels). Viewing the convolutional filters
in equation~\ref{conv} that serve as windowed exponentials as parameters,
the desirable baseline is just one member of a parametric family of models.
This parametric family is known as a ``complex-valued convolutional network.''
We can fit (that is, learn or train) the parameters to the data
via optimization procedures such as stochastic gradient descent
in conjunction with ``backpropagation'' (backpropagation is the chain rule
of Calculus applied to calculate gradients of our recursively composed
operations). For ``supervised learning,''
we optimize according to a specified objective,
usually using the multiscale spectra as inputs to a scheme for classification
or regression, as detailed by~\citet{lecun-bottou-bengio-haffner}, for example.

\begin{remark}
In consonance with the ``best-basis'' approach
of~\citet{coifman-meyer-quake-wickerhauser} and~\citet{saito-coifman},
a potentially more efficient possibility is to restrict
the convolutional filters in equation~\ref{conv} to be windowed exponentials
that are designed completely a priori,
aside from one overall scaling factor per filter,
fitting/learning/training only the scaling factors.
How best to effect this approach is an open question.
\end{remark}

\section{Numerical experiments}
\label{numerical}

The following reports the classification accuracies for the complex-valued
convnets of~\citet{chintala-ranzato-szlam-tian-tygert-zaremba},
the standard real-valued convnets of~\citet{krizhevsky-sutskever-hinton},
and the scattering transforms of~\citet{oyallon-mallat},
on a benchmark data set, ``CIFAR-10,'' from~\citet{krizhevsky}
(CIFAR-10 contains 50,000 images in its training set
and 10,000 images in its testing set; each image falls into one of ten classes,
is full-color, and consists of a $32 \times 32$ grid of pixels):
According to Table~4 of~\cite{oyallon-mallat}, the scattering transforms attain
an error rate of $18\%$ on the test set, after training their classifiers
on the training set. According to Section~3.3
of~\cite{krizhevsky-sutskever-hinton}, a standard real-valued convnet
attains an error rate of $13\%$ on the test set without
the ``local response normalization'' of that Section~3.3,
and attains $11\%$ with the local response normalization.
The complex-valued convnets detailed
in~\citet{chintala-ranzato-szlam-tian-tygert-zaremba}
attain an error rate of $12\%$ on the test set,
at least when using a larger net and training with enough iterations
for the test error to settle down and converge
(for complex-valued convnets, accuracy seems to improve
as the net becomes larger --- for the error rate of $12\%$,
a net eight times the size of that reported in Table~1
of~\citet{chintala-ranzato-szlam-tian-tygert-zaremba} was sufficient,
using the same kernel sizes and other parameter settings as for Table~1).
Augmenting the training images with their mirror images
improved convergence to the reported accuracies. All in all, the extensively
trained real- and complex-valued convnets yielded similar error rates,
which are about a third less than those which scattering transforms attained.
Of course, the fitting/learning/training involved for classification
with the scattering transforms is much less extensive.

\section{Conclusion}
\label{conclusion}

While the above concerns $X_k$,
where $k$ ranges over the integers, extension to analyzing $X_{j,k}$,
where $j$ and $k$ range over the integers, is straightforward
--- the latter could be a ``locally homogeneous random field.''
Also, the infinite range of the integers is far from essential;
implementations on computers obviously use only finite sequences.
Moreover, the above construction is appropriate for processing
any locally stationary stochastic process, not just filtered white noise.
For instance, the construction can enable a multiresolution analysis
of ``regularity'' (or ``smoothness'') that easily distinguishes
between low-pass filtered i.i.d.\ Gaussian noise
and a pulse train or sinusoid with a random phase offset
(for example, $X_k = 1+\sin(\pi (k+J) / 1000)$ for any integer $k$,
where $J$ is an integer drawn uniformly at random from 1, 2, \dots, 2000).
More generally, the construction should enable discriminating
between many interesting classes of stochastic processes,
commensurate with the ability of multiwavelet-based multiresolution analysis
to measure ``regularity,'' ``intermittency,'' distributional characteristics
(say, Gaussian versus Poisson), etc.
Any globally stationary stochastic process --- with or without intermittent
fluctuations --- can be modeled as above as a locally stationary stochastic
process (of course, \citet{bruna-mallat-bacry-muzy} treat the former directly,
to great advantage in the analysis of homogeneous turbulence
and other phenomena from statistical physics).
Every model in the parametric family constituting
the complex-valued convnet calculates relevant features,
windowed spectra of the form in equation~\ref{localized}
and equation~\ref{conv}.
The absolute values in equation~\ref{localized} and equation~\ref{conv}
are the key nonlinearity, a reflection of the local stationarity
--- the local translation-invariance ---
of the process and its relevant features.

\subsection*{Acknowledgments}

We would like to thank Keith Adams, Lubomir Bourdev, Rob Fergus,
Armand Joulin, Manohar Paluri, Christian Puhrsch, Marc'Aurelio Ranzato,
Ben Recht, and Rachel Ward.

\newpage

\bibliography{cc}
\bibliographystyle{apalike}

\end{document}